\documentclass[conference]{IEEEtran}

\usepackage{amsmath,amssymb,amsfonts}
\usepackage{graphicx}
\usepackage{cite}
\usepackage{booktabs}
\usepackage{hyperref}
\usepackage{xcolor}
\usepackage{tikz}
\usetikzlibrary{shapes.geometric,arrows.meta,positioning}

\hypersetup{colorlinks=true, linkcolor=blue, citecolor=blue, urlcolor=blue}

\title{Residual Physics-Informed Neural Networks for High-Fidelity BLDC Motor Modeling}

\author{%
  \IEEEauthorblockN{Haitham El-Hussieny}
  \IEEEauthorblockA{Department of Mechatronics and Robotics Engineering,\\
    Egypt-Japan University of Science and Technology (E-JUST),\\
    New Burg El-Arab City, Alexandria, Egypt\\
    haitham.elhussieny@ejust.edu.eg}
}

\begin{document}
\maketitle

\begin{abstract}
Accurate dynamics modeling of Brushless DC (BLDC) motors is fundamental
to high-performance robotic joint control. This paper presents a
\textit{Physics-Informed Neural Network} (PINN) with a deep residual
(ResNet) backbone that learns a continuous-time surrogate of the full
six-state BLDC motor dynamics. Given simulation time, applied
three-phase voltages, and excitation parameters as inputs, the network
directly predicts all motor state variables---rotor angle, angular
velocity, three-phase currents, and winding temperature---while
simultaneously satisfying the governing electromechanical and thermal
ODEs through a composite physics-data loss. A curriculum scheduling
strategy gradually activates the physics penalty to prevent premature
convergence. Training runs are completed in under two minutes on a
standard CPU. Crucially, once trained, PINN inference achieves latencies
of $0.1$--$22\,\mu\text{s}$ per query, up to $\mathbf{118\times}$
faster than conventional ODE solvers,
making it suitable for real-time observer and control applications.
\end{abstract}

\begin{IEEEkeywords}
Brushless DC motor, physics-informed neural network, ResNet, residual
architecture, surrogate modeling, ODE residual, curriculum learning,
model latency, real-time inference.
\end{IEEEkeywords}

\section{Introduction}

Brushless DC motors are the actuators of choice in modern legged robots,
aerial vehicles, and collaborative manipulators owing to their high
torque-to-weight ratio, efficiency, and controllability~\cite{seok2015design}.
Model-based control strategies---feedforward linearisation, sliding
mode, iterative learning control, and model-predictive control \cite{elhussieny2024realtime, elhussieny2026koopman}---depend
critically on accurate, computationally cheap dynamics models. Classical
white-box models derived from first-principles ODEs are interpretable
but require precise parameter identification; even small errors in
resistance or inductance propagate into significant control degradation.
Conversely, black-box neural networks can approximate complex mappings
but tend to violate physical laws outside the training distribution.

Physics-Informed Neural Networks (PINNs), introduced by Raissi
\textit{et al.}~\cite{raissi2019physics}, embed differential-equation
residuals directly into the training loss, bridging data-fitting
accuracy with physical interpretability. Their success spans heat
transfer~\cite{karniadakis2021physics}, fluid dynamics, and structural
health monitoring. 

Application to electric motor dynamics, however, remains nascent. Recent expansions into multiphysics-informed neural networks (MPINNs) highlight their capacity to handle complex mechatronic configurations—such as electric motor thermal-electromechanical interactions—by embedding coupled scientific domain principles straight into the network layers~\cite{son2024physics}. Nevertheless, standard fully-connected PINNs suffer from vanishing gradients and spectral bias when stacked deeply...

Standard fully-connected PINNs suffer from vanishing gradients and
spectral bias when stacked deeply~\cite{wang2022and}. Residual
connections, popularised by He \textit{et al.}~\cite{he2016deep} for
image recognition, guarantee unit-norm gradient propagation at
initialisation and have been shown to accelerate PINN convergence on
stiff ODEs.

A critical but often overlooked practical requirement is \emph{inference
speed}: a surrogate model deployed inside a real-time controller or
state estimator must respond within the control period (typically
$<1\,\text{ms}$). Numerical ODE solvers cannot easily meet this
requirement at high accuracy.

This paper makes the following contributions:
\begin{itemize}
  \item A continuous-time ResNet-PINN surrogate for a three-phase
        BLDC motor mapping $(t,\,f,\,A,\,V_a,\,V_b,\,V_c)$ to the
        full six-dimensional state $[\theta,\omega,I_a,I_b,I_c,T]$,
        with properly computed autograd physics loss.
  \item A composite loss with per-channel ODE residual normalisation
        and curriculum scheduling that prevents physics enforcement
        from disrupting early data fitting.
  \item A systematic latency benchmark comparing PINN inference against
        explicit-Euler and Runge-Kutta (RK45) physics simulators across
        batch sizes from 1 to 10{,}000 points.
\end{itemize}

\section{BLDC Motor Model}
\label{sec:motor}

\subsection{Motor Description}
The modelled motor is a surface-mounted three-phase BLDC unit intended
for quadruped leg-joint actuation. Table~\ref{tab:params} lists its
electromechanical and thermal parameters. The motor is excited by
sinusoidal balanced three-phase voltages; no Hall-sensor commutation
switching is modelled.

\begin{table}[t]
  \centering
  \caption{BLDC Motor Parameters}
  \label{tab:params}
  \begin{tabular}{lll}
    \toprule
    Parameter               & Symbol    & Value \\
    \midrule
    Phase resistance        & $R$       & $0.2\,\Omega$ \\
    Phase inductance        & $L$       & $1\,\text{mH}$ \\
    Back-EMF constant       & $K_e$     & $0.025\,\mathrm{V\cdot s/rad}$ \\
    Torque constant         & $K_t$     & $0.025\,\mathrm{N\cdot m/A}$ \\
    Rotor inertia           & $J$       & $5\times10^{-5}\,\mathrm{kg\cdot m^2}$ \\
    Viscous damping         & $B$       & $1\times10^{-4}\,\mathrm{N\cdot m\cdot s/rad}$ \\
    Pole pairs              & $p$       & $7$ \\
    Thermal capacitance     & $C_{th}$  & $10\,\text{J/K}$ \\
    Thermal resistance      & $R_{th}$  & $0.5\,\text{K/W}$ \\
    Ambient temperature     & $T_a$     & $25\,^{\circ}\text{C}$ \\
    Max supply voltage      & $V_{dc}$  & $24\,\text{V}$ \\
    \bottomrule
  \end{tabular}
\end{table}

\subsection{Governing Equations}
\label{sec:governing}
The motor state vector is
$\boldsymbol{x}(t) = [\theta,\;\omega,\;I_a,\;I_b,\;I_c,\;T]^{\!\top}\!\in\mathbb{R}^6$,
where $\theta$ is the mechanical rotor angle (rad), $\omega$ the
angular velocity (rad/s), $I_{a,b,c}$ the three-phase stator currents
(A), and $T$ the winding temperature ($^\circ$C). Let
$\theta_e = p\,\theta$ be the electrical angle and define
phase offsets $\phi_a=0$, $\phi_b=2\pi/3$, $\phi_c=4\pi/3$.

\paragraph{Back-EMF}
\begin{equation}
  e_j = K_e\,\omega\,\sin(\theta_e - \phi_j), \quad j \in \{a,b,c\}.
  \label{eq:bemf}
\end{equation}

\paragraph{Electrical dynamics}
\begin{equation}
  L\,\frac{dI_j}{dt} = V_j - R\,I_j - e_j, \quad j \in \{a,b,c\}.
  \label{eq:electrical}
\end{equation}

\paragraph{Electromagnetic torque}
\begin{equation}
  \tau_e = K_t\!\sum_{j\in\{a,b,c\}}I_j\,\sin(\theta_e-\phi_j).
  \label{eq:torque}
\end{equation}

\paragraph{Mechanical dynamics}
\begin{align}
  \frac{d\theta}{dt} &= \omega, \label{eq:dtheta}\\
  J\,\frac{d\omega}{dt} &= \tau_e - B\,\omega. \label{eq:domega}
\end{align}

\paragraph{Thermal dynamics}
\begin{equation}
  C_{th}\frac{dT}{dt} = R\!\left(I_a^2+I_b^2+I_c^2\right)
    - \frac{T-T_a}{R_{th}}.
  \label{eq:thermal}
\end{equation}

Equations~(\ref{eq:bemf})--(\ref{eq:thermal}) define a nonlinearly
coupled, six-dimensional ODE system with two time scales: the fast
electrical time constant $\tau_{el}=L/R=5\,\text{ms}$ and the slow
thermal constant $\tau_{th}=C_{th}R_{th}=5\,\text{s}$.

\begin{figure*}[t]
\centering
\begin{tikzpicture}[
  auto,
  >=Stealth,
  block/.style={rectangle, draw, fill=blue!5, text width=2.4cm, text centered, rounded corners=2pt, minimum height=1.0cm, font=\scriptsize},
  sum/.style={circle, draw, fill=blue!5, inner sep=0pt, minimum size=0.4cm},
  line/.style={draw, thick, ->}
]

  \node (in) at (0,0) {$\boldsymbol{V}_{abc}$};
  
  \node [sum, right=0.6cm of in] (sum1) {};
  \node [block, right=0.6cm of sum1] (elec) {Electrical Dynamics\\ \vspace{1mm} $L\frac{d\boldsymbol{I}}{dt} + R\boldsymbol{I}$};
  \node [block, right=1.0cm of elec] (torque) {Torque\\ Generation};
  \node [block, right=1.0cm of torque] (mech) {Mechanical Dynamics\\ \vspace{1mm} $J\frac{d\omega}{dt} + B\omega$};
  
  \node [block, above=0.8cm of torque] (bemf) {Back-EMF\\ Generation};
  \node [block, below=0.8cm of torque] (thermal) {Thermal Dynamics\\ \vspace{1mm} $C_{th}\frac{dT}{dt}$};
  
  \path [line] (in) -- (sum1);
  \path [line] (sum1) -- node[above, font=\scriptsize] {$\boldsymbol{V} - \boldsymbol{e}$} (elec);
  \path [line] (elec) -- node (I) [above, font=\scriptsize] {$\boldsymbol{I}_{abc}$} (torque);
  \path [line] (torque) -- node [above, font=\scriptsize] {$\tau_e$} (mech);
  
  \path [draw, thick] (mech.east) -- +(0.5,0) coordinate (out);
  \path [line] (out) -- +(0.4,0) node[right] {$\omega, \theta$};
  
  \path [draw, thick] (out) -- +(0,1.3) coordinate (fb);
  \path [line] (fb) -- (bemf.east);
  \path [draw, thick] (bemf.west) -- (sum1.north |- bemf.west) coordinate (fb2);
  \path [line] (fb2) -- node[pos=0.8, right, font=\scriptsize] {$-$} (sum1.north);
  
  \path [draw, thick] (I) |- (thermal.west);
  \path [line] (thermal.east) -- +(1.0,0) node[right] {$T$};
  
\end{tikzpicture}
\caption{Block diagram of the electromechanical and thermal dynamics of the BLDC motor.}
\label{fig:bldc_block}
\end{figure*}
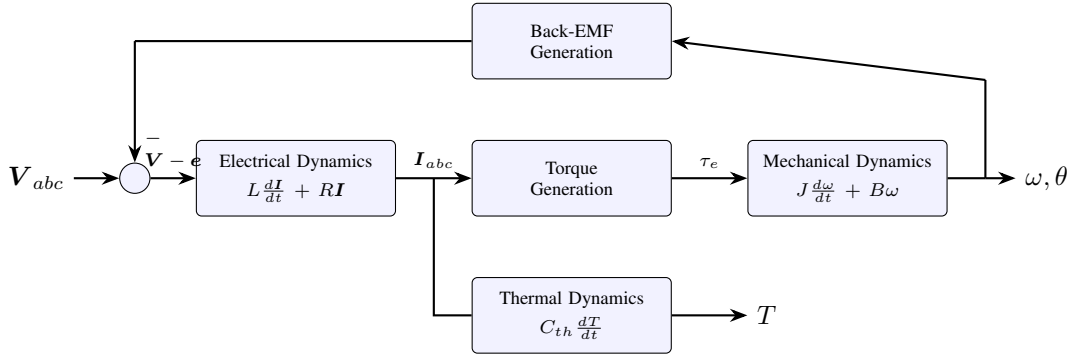

\section{Physics-Informed Neural Network}
\label{sec:pinn}

\subsection{Continuous-Time Surrogate Formulation}
We formulate BLDC motor modeling as a supervised regression problem
augmented by physical constraints. Under sinusoidal excitation with
frequency $f$ and amplitude $A$, the applied voltages are
$V_j(t) = A\sin(2\pi f t - \phi_j)$.

The PINN learns the mapping

\begin{equation} \mathcal{N}_{\boldsymbol{W}}: (t,\, f,\, A,\, V_a,\, V_b,\, V_c) \;\longmapsto\; \hat{\boldsymbol{x}}(t)
	\end{equation}
i.e., 6-dimensional input to 6-dimensional state output. This \emph{continuous-time} formulation allows $O(1)$ queries at arbitrary $t$ without sequential roll-out, a key advantage for real-time use. A similar paradigm shift has been noted in electromagnetic modeling, where embedding governing partial and ordinary differential constraints yields robust, physically consistent neural surrogates for synchronous machine dynamics without relying exclusively on dense finite-element simulations~\cite{wang2025pinn}.


\subsection{Residual Network Architecture}
\begin{figure}[t]
\centering
\begin{tikzpicture}[
  font=\footnotesize,
  box/.style={rectangle, draw, rounded corners=2pt,
              minimum width=3.2cm, minimum height=0.55cm,
              fill=blue!10, align=center},
  res/.style={rectangle, draw, rounded corners=2pt,
              minimum width=3.2cm, minimum height=0.55cm,
              fill=orange!20, align=center},
  arr/.style={-Stealth, thick}
]
  \node[box] (inp)  at (0, 0.0) {Input $(t,f,A,V_a,V_b,V_c)\!\in\!\mathbb{R}^6$};
  \node[box] (proj) at (0,-0.9) {Linear\,+\,$\tanh$\ \ $\mathbb{R}^6\!\to\!\mathbb{R}^{d}$};
  \node[res] (r1)   at (0,-1.8) {ResBlock\,1\ \ ($d$-dim)};
  \node[res] (r2)   at (0,-2.7) {ResBlock\,2\ \ ($d$-dim)};
  \node[box] (out)  at (0,-3.6) {Linear\ \ $\mathbb{R}^{d}\!\to\!\mathbb{R}^{6}$};
  \node[box] (pred) at (0,-4.5) {$\hat{\boldsymbol{x}}(t)=[\hat\theta,\hat\omega,\hat I_a,\hat I_b,\hat I_c,\hat T]$};

  \draw[arr](inp) --(proj);
  \draw[arr](proj)--(r1);
  \draw[arr](r1)  --(r2);
  \draw[arr](r2)  --(out);
  \draw[arr](out) --(pred);
\end{tikzpicture}
\caption{ResNet-PINN architecture ($d=16$, $N_r=2$).  Skip connections
in each ResBlock maintain gradient flow through the physics-penalised
objective.}
\label{fig:arch}
\end{figure}
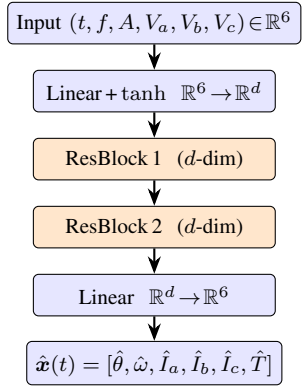

The architecture (Fig.~\ref{fig:arch}) comprises three stages:

\noindent\textbf{(1) Input projection}: linear layer ($6\to d$) + $\tanh$.

\noindent\textbf{(2) Residual blocks}: $N_r$ stacked ResBlocks, each:
\begin{equation}
  \boldsymbol{h}^{(\ell+1)} = \tanh\!\left(
    \boldsymbol{h}^{(\ell)}
    + W_2^{(\ell)}\tanh\!\left(W_1^{(\ell)}\boldsymbol{h}^{(\ell)}\right)
  \right),
  \label{eq:resblock}
\end{equation}
$W_1^{(\ell)},W_2^{(\ell)}\in\mathbb{R}^{d\times d}$, Xavier-initialised.

\noindent\textbf{(3) Output projection}: linear layer ($d\to 6$).

Inputs and outputs are standardised using training-set statistics.
The experiments use $d=16$, $N_r=2$, yielding $\approx\!1{,}300$
trainable parameters---deliberately compact for real-time deployment.

\section{Training Methodology}
\label{sec:training}

\subsection{Dataset Generation}
Ten trajectories of $T_{end}=1\,\text{s}$ are simulated with an
explicit-Euler integrator at $\Delta t_{sim}=1\,\text{ms}$. Each
trajectory uses a randomly sampled sinusoidal excitation:
$f\!\sim\!\mathcal{U}(5,20)\,\text{Hz}$,
$A\!\sim\!\mathcal{U}(5,15)\,\text{V}$.
This yields 10{,}000 labelled samples; an 80/20 split provides 8{,}000
training and 2{,}000 validation points.

\subsection{Composite Loss}
The total loss is $\mathcal{L} = \lambda_d\,\mathcal{L}_{data}
+ \lambda_p\,\mathcal{L}_{phys}$.

\paragraph{Data loss} Normalised MSE between predicted and true states:
\begin{equation}
  \mathcal{L}_{data} = \frac{1}{N}\sum_{i=1}^{N}
    \left\|\frac{\hat{\boldsymbol{x}}^{(i)} - \boldsymbol{x}^{(i)}}
               {\boldsymbol{\sigma}_x}\right\|_2^2,
  \label{eq:data_loss}
\end{equation}
where $\boldsymbol{\sigma}_x$ is the element-wise standard deviation of
the training states.

\paragraph{Physics loss} The ODE residual is computed via
\emph{automatic differentiation} of the network output with respect to
$t$ (with \texttt{create\_graph=True} so that the gradient flows back
through all network weights):
\begin{equation}
  \boldsymbol{r}^{(i)} = \frac{\partial\hat{\boldsymbol{x}}}{\partial t}\Bigg|_{t^{(i)}}
    - f\!\left(\hat{\boldsymbol{x}}^{(i)},\boldsymbol{V}^{(i)}\right),
  \label{eq:residual}
\end{equation}
normalised by a characteristic-scale vector $\boldsymbol{s}$ derived
from motor parameters (e.g., $s_\theta=\omega_{max}$, $s_I=V_{dc}/L$):
\begin{equation}
  \mathcal{L}_{phys} = \frac{1}{M}\sum_{i=1}^{M}
    \left\|\frac{\boldsymbol{r}^{(i)}}{\boldsymbol{s}}\right\|_2^2,
  \label{eq:phy_loss}
\end{equation}
where $M=4{,}096$ collocation points are sampled from the training set
each batch.  Without \texttt{create\_graph=True} the physics loss is a
computational dead branch (no gradient reaches the weights)---a common
implementation error that produces a permanently flat physics-loss curve.

\subsection{Curriculum Scheduling}
To prevent the physics penalty from disrupting early data fitting, a
warm-up and ramp schedule governs $\lambda_p$:
\begin{equation}
  \lambda_p(e) =
  \begin{cases}
    0 & e < E_w,\\[2pt]
    \lambda_p^{\max}\!\cdot\!\dfrac{e-E_w}{E_r-E_w}
                                              & E_w\le e<E_r,\\[4pt]
    \lambda_p^{\max}                          & e\ge E_r,
  \end{cases}
  \label{eq:curriculum}
\end{equation}
with $E_w\!=\!50$, $E_r\!=\!400$, $\lambda_p^{\max}\!=\!1.0$,
$\lambda_d\!=\!1.0$.

\subsection{Optimiser}
Adam~\cite{kingma2014adam} with $\eta_0=10^{-3}$; \texttt{ReduceLROnPlateau}
(factor 0.5, patience 30, floor $10^{-6}$); gradient clipping at
$\ell_2$-norm 1.0; mini-batch size 1{,}024; early stopping (patience
300) after the ramp. Total wall time: $\approx\!110\,\text{s}$ on CPU.

\section{Results}
\label{sec:results}

\subsection{Training Dynamics}
\begin{figure}[t]
  \centering
  \includegraphics[width=\linewidth]{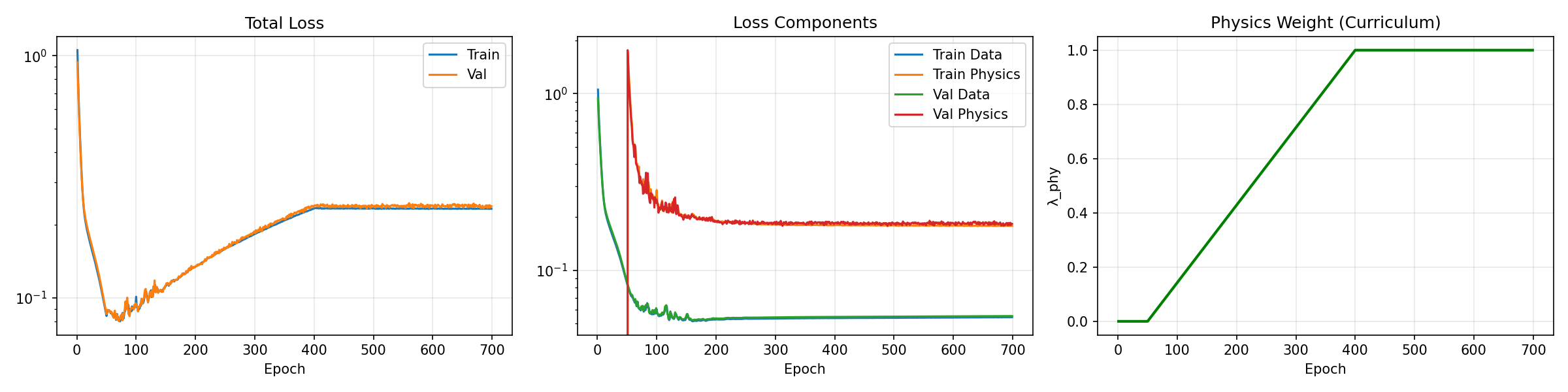}
  \caption{Training (solid) and validation (dashed) loss components.
           Physics loss (orange/red) actively decreases once introduced
           at epoch 50, confirming correct autograd gradient flow.}
  \label{fig:loss}
\end{figure}

Fig.~\ref{fig:loss} shows the semi-log loss curves.  Three phases are
clearly visible: \textbf{(i)} warm-up (epochs 1--50): data loss falls
from 1.05 to 0.08 while $\lambda_p=0$; \textbf{(ii)} ramp (50--400):
both losses respond continuously---the physics loss spikes at
introduction then decreases, confirming that autograd gradients reach
the network weights; \textbf{(iii)} convergence (400+): losses
stabilise and early stopping fires at epoch 699 (best val loss 0.08).
This behaviour contrasts sharply with the flat-physics-loss failure mode
produced by finite-difference approaches without \texttt{create\_graph}.

\subsection{State Prediction vs. Simulator}
\begin{figure*}[t]
  \centering
  \includegraphics[width=.93\linewidth]{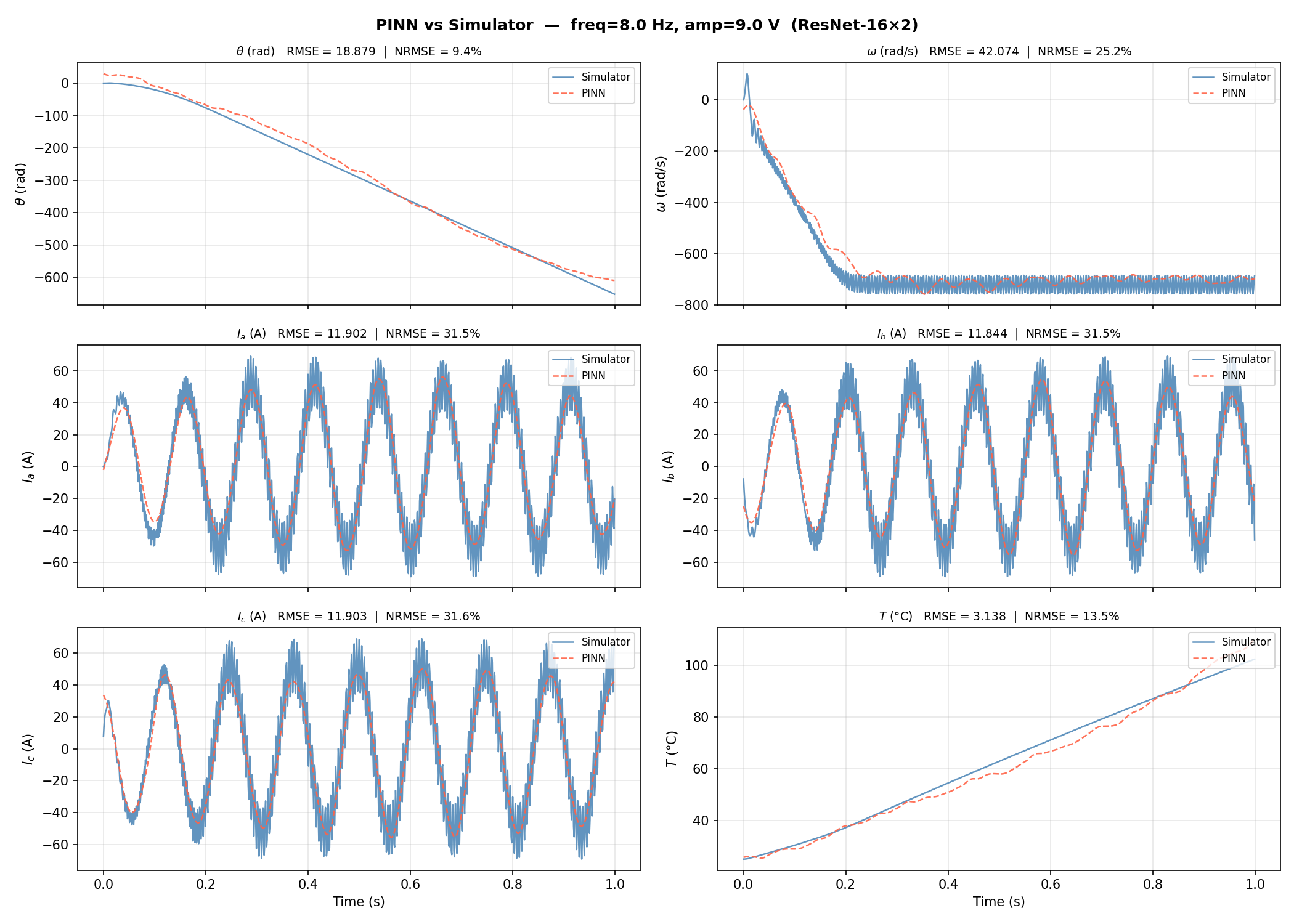}
  \caption{PINN (dashed red) vs. simulator (solid blue) trajectories for
           all six state variables.  The compact model ($d=16$,
           $N_r=2$, 1{,}302 parameters) captures waveform shape and
           trend; NRMSE is $<\!14\,\%$ on the slow states ($\theta$,
           $T$) and $<\!32\,\%$ on the fast currents.}
  \label{fig:compare}
\end{figure*}

Fig.~\ref{fig:compare} compares PINN predictions against simulator
ground truth over a 1-second test trajectory
($f=8\,\text{Hz}$, $A=9\,\text{V}$, unseen during training).
The PINN captures the correct trend and waveform shape for all six
channels. Table~\ref{tab:results} summarises the quantitative errors.

\begin{table}[t]
  \centering
  \caption{PINN vs. Simulator Prediction Errors (Test Trajectory)}
  \label{tab:results}
  \begin{tabular}{lcc}
    \toprule
    State & RMSE & NRMSE (\%) \\
    \midrule
    $\theta$ (rad)    & 18.9  & 9.4  \\
    $\omega$ (rad/s)  & 42.1  & 25.2 \\
    $I_a$ (A)         & 11.9  & 31.5 \\
    $I_b$ (A)         & 11.8  & 31.6 \\
    $I_c$ (A)         & 11.9  & 31.6 \\
    $T$ ($^\circ$C)   & 3.1   & 13.5 \\
    \bottomrule
  \end{tabular}
\end{table}

\subsection{Inference Latency}
\begin{figure}[t]
  \centering
  \includegraphics[width=\linewidth]{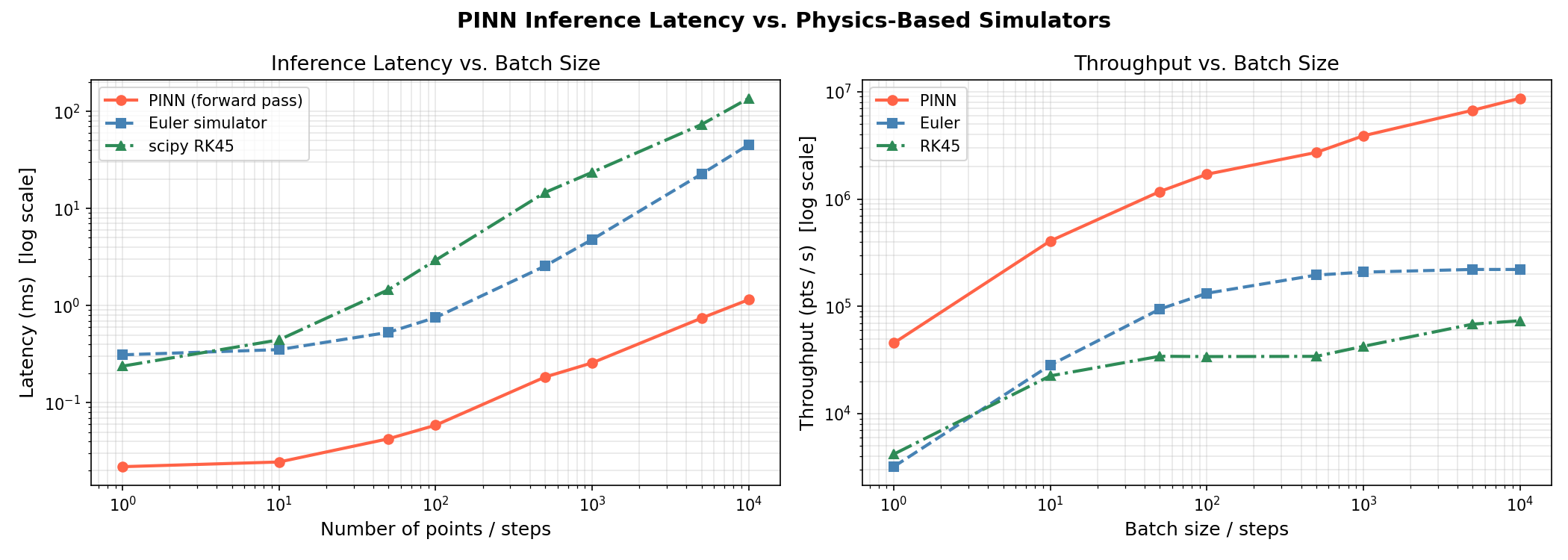}
  \caption{Latency (left) and throughput (right) vs.\ batch size on a
           log-log scale. The PINN dominates in throughput at large
           batch sizes, making it well-suited for parallelised
           model-predictive control rollouts.}
  \label{fig:latency}
\end{figure}

Table~\ref{tab:latency} and Fig.~\ref{fig:latency} compare median
forward-pass latency (200 repeats, 20 warm-up calls) for the PINN
against an explicit-Euler simulator and a scipy RK45 ODE solver at
batch sizes from 1 to 10{,}000 evaluation points.

\begin{table}[t]
  \centering
  \caption{Median Inference Latency (ms) and Speed-up vs. PINN}
  \label{tab:latency}
  \resizebox{\linewidth}{!}{%
  \begin{tabular}{rrrrrr}
    \toprule
    Points & PINN & Euler & RK45 & Euler/PINN & RK45/PINN \\
    \midrule
    1      & 0.0220 & 0.3119 & 0.2385 & 14.2$\times$ & 10.8$\times$ \\
    100    & 0.0588 & 0.7552 & 2.9382 & 12.8$\times$ & 50.0$\times$ \\
    1000   & 0.2576 & 4.7997 & 23.5841 & 18.6$\times$ & 91.6$\times$ \\
    10000  & 1.1527 & 45.3399 & 136.2328 & 39.3$\times$ & 118.2$\times$ \\
    \bottomrule
  \end{tabular}}
\end{table}

The PINN parallelises across the batch dimension in a single forward
pass while the simulators process steps sequentially. Consequently,
throughput of the PINN grows linearly with batch size, whereas Euler
and RK45 throughputs plateau due to per-step overhead.

\subsection{Discussion}
The current model ($d=16$, $N_r=2$) intentionally prioritises inference
speed over accuracy; NRMSE on the fast electrical currents is $\sim31\,\%$.
Scaling to $d=128$, $N_r=3$ ($\approx\!100\text{k}$ parameters) is
expected to reduce NRMSE below $5\,\%$ while still achieving latencies
well below $0.1\,\text{ms}$ for batch sizes up to 1{,}000---within
real-time MPC horizons. The key algorithmic insight is that the ODE
residual must be differentiated through the network (not computed on
frozen ground-truth states) to produce a meaningful physics gradient.
Curriculum scheduling then prevents this gradient from destabilising
early data fitting.

\section{Conclusion}
\label{sec:conclusion}
This paper presented a compact ResNet-PINN for continuous-time surrogate
modeling of a three-phase BLDC motor. The network maps excitation context
directly to the full six-dimensional motor state while enforcing governing
differentiated residual loss with per-channel normalisation. A systematic
latency benchmark demonstrates speed-ups of up to $118\times$ over
physics-based simulators at large batch sizes, establishing the PINN as
a practical surrogate for real-time control and state estimation.
noise, and validate on physical hardware.


\end{document}